\documentclass[11pt]{article}

\usepackage{graphicx}
\usepackage{amssymb}
\usepackage{amsmath}
\usepackage{bm} %maths

\usepackage[latin1]{inputenc}  

\usepackage{color}

\renewcommand{\d}{\mathrm{d}}
\newcommand{\grad}{\mathbf{grad}}
\newcommand{\dive}{\mathrm{div}}

\newcommand{\ex}{\mathrm{\textbf{e}}_1}
\newcommand{\ey}{\mathrm{\textbf{e}}_2}

\newcommand{\vecx}{\mathrm{\textbf{x}}}
\newcommand{\vecX}{\mathrm{\textbf{X}}}
\newcommand{\vecn}{\mathrm{\textbf{n}}}
\newcommand{\vectau}{\bm{\tau}}
\newcommand{\vecnu}{\bm{\nu}}

\begin{document}

\title{Analytical shape determination of fiber-like objects with Virtual Image Correlation}

\author{Beno{\^i}t Semin, Marc Fran\c cois*, Harold Auradou}

\date{\today}

\maketitle

Lab. FAST, Univ. Paris-Sud 11, Bat. 502, Orsay F-91405, France

\small
* Corresponding author - email: marc.francois@u-psud.fr
\normalsize 

\begin{abstract}
%% Text of abstract

This paper reports a method allowing for the determination of the shape of deformed fiber-like objects. Compared to existing methods, it provides analytical results including the local slope and curvature which are of first importance, for instance, in beam mechanics. The presented VIC (Virtual Image Correlation) method consists in looking for the best correlation between the image of the fiber-like object and a virtual beam image, using an algorithm close to the Digital Image Correlation method developed in experimental solid mechanics. The computation only involves the part of the image in the vicinity of the fiber: the method is thus insensitive to the picture background and the computational cost remains low. Two examples are reported: the first proves the precision of the method, the second its ability to identify a complex shape with multiple loops.

\end{abstract}

keywords : analytical shape, curve detection, virtual image correlation, fiber, filament, loop

%% main text
%---------------------------------------------------------------------
\section{Introduction}
\label{sec:intro}
% Urbi et orbi
The determination of the shape and position of elongated objects such as hair \cite{Robbins2002}, pulp fibers \cite{Forgacs1959}, needles \cite{Okazawa2006}, biological filaments \cite{Shin2004,Diluzio2005} or abiological objects \cite{Dreyfus2005,Garstecki2009} is of interest in various fields of research. 
% Tordu
These objects may be bent by internal or external forces induced, for instance, by their own weight, by flowing fluids or by the interaction with solid surfaces.
% State of the art
Image processing offers a set of techniques and algorithms that may be used for the detection of such features.
The basic technique consists in thresholding the image and skeletonizing it.
An other possibility is to follow the line of maximal intensity (called "ridge") using the eigenvector corresponding to the highest eigenvalue of the Hessian of the (smoothed) image \cite{Aylward2002} or by using the minimal path method \cite{Deschamps2001,Mueller2008}.
Among the various families of techniques, a class of method is based on the Hough or the Radon Transform \cite{Xu1990,Lam1993,Toft1996} which consist in transforming the images in such a way that segments become points, easily detected by image thresholding.
Finally, the level set method also provides an efficient measurement of contours \cite{Caselles1997}.
% Pas glop
Yet, while these methods allow one to determine the shape of a curvilinear object, they do not allow one to estimate precisely its local curvature: this is however a key parameter which may be directly related to its mechanical state by the beam theory \cite{Timoschenko1924}.
%Yet, while these methods may be used to extract the shape of the object, they are not made to give an accurate estimation of the local curvature while the mechanical state depends on it by the beam theory \cite{Timoschenko1924}.

% VIC pour les nuls
In the present method, the fiber shape is given by the optimal correlation between its physical image and a virtual beam.
The latter has a mean line defined from a series expansion and a gray level which smoothly decreases from the mean line to the borders.
% This one has a mean line expressed as a series expansion of its curvatures and a smoothly decreasing lightness along its radius.
% its width is slightly greater than the physical one.
% VIC pour les nuls
%In the proposed method, the beam's mean line curvatures are defined by the coefficients of a series (Fourier or Legendre in the proposed examples). Then the researched mean line is supposed infinitely smooth, allowing an easy and robust determination of the slopes and curvatures. The virtual beam is defined from this mean line; it has a user-defined thickness, close to the physical beam width. Its lightness decreases from white at the mean line to black at the sides (in retained examples, the physical beams are light gray on dark background).  The beam shape is defined as the one that provides the best correlation between the physical image and the virtual beam.
The correlation only involves the definition domain of the virtual beam (close to the physical one) that represents in general a much smaller number of pixels than the complete image. 
%For this reason, the method is fully insensitive to far artefacts (see the first example) and does not require huge computation time. 
The correlation algorithm uses recent developments of the Digital Image Correlation techniques (DIC) and its application to mechanics \cite{Hild2006}. This operation does not require any light intensity thresholding.
% and the best adjustment give the mean line of the physical object. 

% Plan du papier
The characteristics of the virtual beam are introduced in Sec.~\ref{sec:virtual}.
The mathematical technique used for correlating the virtual beam onto the raw experimental image is proposed in Sec.~\ref{sec:adj}.
Sec.~\ref{sec:init} details the determination of the initial parameters required by the method.
Two practical applications of the Virtual Image Correlation (VIC) technique are reported in Sec.~\ref{sec:beam} and \ref{sec:fiber}.
The first one consists of a straight cantilever elastic beam bending under its own weight: it is shown that the measurement is consistent with the theoretical result given by the beam theory.
The second uses an experimental low resolution and noisy picture of a fiber transported (and curved) by a flow in a fracture. The complex shape is recovered, demonstrating the robustness of the method with regards to loops, noise, luminance and contrast variations.
Further developments of the VIC method are finally discussed in Sec.~\ref{sec:conclusion}.\\

%---------------------------------------------------------------------
\section{Parameterization of the virtual image}
\label{sec:virtual}

\begin{figure}[htbp]
\begin{center}
\includegraphics[scale=0.4]{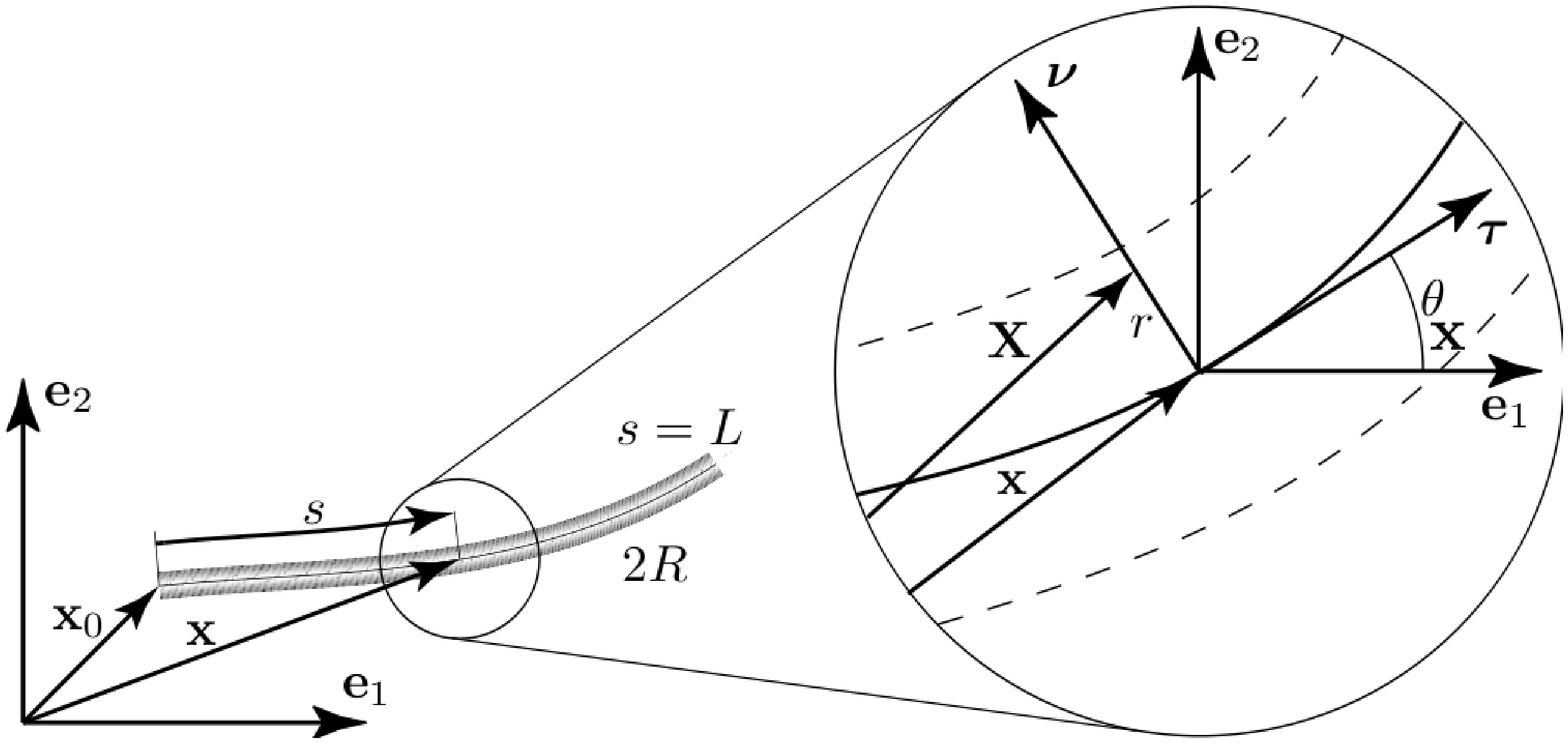}
\caption{The virtual beam image and its coordinates system}
\label{sketch}
\end{center}
\end{figure}

The virtual image G consists of the virtual beam (in the sense of a plane curvilinear object in mechanics) with a length $L$ and a width $2R$.
Any point $\vecX$ of the beam is parameterized by its curvilinear abscissa $s\in[0,L]$ along the mean line and its transverse distance $r\in[-R,R]$ from it (see Fig.~\ref{sketch}); the image G is not defined outside the beam's definition domain $D_g$.
The local curvature is $\gamma(s)$, the angle is $\theta(s)$, with $\theta_0 = \theta(s=0)$.
The cartesian reference frame is $(\ex,\ey)$, the tangent and normal vectors are respectively $(\vectau, \vecnu)$ and a point of the mean line is referred to as $\vecx$ (Eq.~\ref{defX} and \ref{defx}).
There is no overlap, \emph{i.e.} the points $\vecX$ are uniquely defined, since the local radius of curvature $1/|\gamma|$ is greater than the beam thickness R. 
%
%The tangent and normal vectors to the mean line are given by:
%\begin{eqnarray}
%    \vectau &=& \cos(\theta)\ex + \sin(\theta)\ey\\
%    \vecnu &=&\ez \times \vectau,
%\end{eqnarray}
%\begin{eqnarray}
%    \theta &=& \theta_0 + \int_0^{s} \gamma \d \xi\label{deftheta}
%\end{eqnarray}
%
\begin{eqnarray}
    \vecX &=& \vecx + r \vecnu \label{defX}\\
    \vecx &=& \vecx_0 + \int_0^{s} \vectau(\xi) \d \xi\label{defx}
\end{eqnarray}
%The shape of the object will be identified by image correlation between the physical object image and the virtual beam. 
In order to give a finite dimension to the problem, the curvature $\gamma(s)$ is described by a truncated series:
\begin{equation}
    \gamma(s) = \sum_{n=0}^{N} A_n \tilde{\gamma}_n(\tilde{s})\label{defan},
\end{equation}
where $A_n$ are the coefficients of the series, $N$ its order, $\tilde{\gamma}_n(\tilde{s})$ the dimensionless basis functions and $\tilde{s}=s/L\in[0,1]$ the reduced curvilinear abscissa.
By integration, the angles are given by:
\begin{equation}
    \theta(s) = \theta_0 + L \sum_{n=0}^{N} A_n \tilde{\theta}_n(\tilde{s})\label{thetaa}
\end{equation}
in which
\begin{equation}
    \tilde{\theta}_n(\tilde{s}) = \int_0^{\tilde{s}} \tilde{\gamma}_n(\tilde{\xi}) \d \tilde{\xi}\label{defthetan}.
\end{equation}
Both $\tilde{\gamma}_n$ and $\tilde{\theta}_n$ depend only upon the choice of the series. They are estimated and stored prior to any computation. The set of parameters which must be determined are the magnitudes $A_n$ and $(\vecx_0,\theta_0$).

% La fonction g
The value of the gray levels $g(\vecX)$ in the virtual image is chosen to be similar to that in the physical objects; it consists of a symmetrical function of the radius $g(\vecX) = l(r)$ that continuously decreases from the mean line to the border:
\begin{equation}
    2l(r) = 1+\cos\left( \frac{\pi r}{R} \right)\label{graylevel}.
%   2l'(r)&=&-\frac{\pi}{R}\sin\left(\frac{\pi r}{R} \right).
\end{equation}
The present method does not require that the virtual image be closely similar to the physical one. Both the width and the gray levels of the virtual beam may be very different from the physical ones; this will be illustrated by the examples.
\section{Adjustment of the virtual beam to the physical picture}
\label{sec:adj}

As shown in the previous section, for a given set of function $\tilde{\gamma}_n$, the shape of the virtual image G is fully described by the set of parameters $\mathbf{V} = \{\vecx_0,\theta_0,A_n\}$ (a pseudo-vector of dimension $N+4$ whose components are referred as $V_k$ hereafter in this section). In this section, we describe the method developed to find the optimal set  $\mathbf{V}^f$, such that the beam image G falls best onto the image of the physical object that is contained in image F. Following \cite{Hild2006}, this is achieved by the minimization of the function $\Phi(V_k)$:
\begin{equation}
    \Phi = \iint_{D_g} \left( f-g \right)^2 \d S \label{defphi}
\end{equation}
in which $f$ is the luminance of the physical image F and $\d S=(1-\gamma r) \d r \d s$ is the surface element. The domain $D_g$ is fully contained inside F so that $f$ is defined at any points. The optimal set $\mathbf{V}^f$ corresponds to the minimum of $\Phi$ then the condition $\d \Phi=(\partial \Phi / \partial V_k) \d V_k=0$ must be fulfilled for any variation $\d V_k$ around $V_k^f$. Using Eq.~(\ref{defphi}) together with $\partial g/\partial V_k = \grad(g).\partial \vecX/\partial V_k$, this condition leads to:
\begin{eqnarray}
\oint_{\partial D_g} \left( (f-g)^2 \, \vecn.\frac{\partial \vecX}{\partial V_k}\d V_k \right) \d l 
- 2 \iint_{D_g} \left( f-g \right)\left(\grad(g).\frac{\partial \vecX}{\partial V_k}  \d V_k \right) \d S = 0,\label{dphieqzero}
\end{eqnarray}
where $\partial D_g$ is the boundary of the domain $D_g$ \emph{i.e.} the external boundary of the virtual beam, $\d l$ a differential line element and $\vecn$ a vector normal to the boundary and pointing outwards. Supposing that the iterative process is close to the solution, and that $R$ is slightly greater than the width of the physical object, this boundary $\partial D_g$ is located in the background of the physical image (we neglect the boundary sides at $s=0$ and $s=L$ as $R<<L$). Then, assuming that the background is uniform, $f|_{\partial D_g}$ (which represents the value of $f$ along the boundary $\partial D_g$) is constant. Furthermore, with the retained definition of $g$ (Eq.~\ref{graylevel}), $g|_{\partial D_g}=0$. Using the divergence theorem, we have:
\begin{eqnarray}
    \oint_{\partial D_g} (f-g)^2 \, \vecn.\frac{\partial \vecX}{\partial V_k}\d V_k\quad \d l =
     (f|_{\partial D_g})^2 \frac{\partial}{\partial V_k} \left( \iint_{D_g}  \dive (\vecX) \d S \right) \d V_k\label{intcontour}
\end{eqnarray}
As $\dive (\vecX)=2$, the surface integral in Eq.~(\ref{intcontour}) is constant (as $S=2RL$) and its derivative with respect to $V_k$ equals zero. 
Thus, assuming henceforth that the virtual beam boundary encloses the physical one and that the background of $f$ is uniform, Eq.~\ref{dphieqzero} reduces to:
%
%(even at the first steps of the procedure, this done thanks to the approximate solution determination in %Sec.~\ref{sec:init}) 
%
%\textbf{Vieille autre demo
%Then we have:
%\begin{eqnarray}
%   \d \Phi &\simeq& <f|_{\partial D_g}>^2 \d V_k
%   \int_{\partial D_g} \, \vecn.\frac{\partial \vecX}{\partial V_k}\quad \d l\nonumber\\
%   &-& 2 \iint_{D_g} \left( f-g \right)\left( \grad(g).\frac{\partial \vecX}{\partial V_k}  \d V_k \right)\quad \d S,\label{demomarc}
%\end{eqnarray}
%Excepting the boundary part of the beam, at each curvilinear abscissa $s$ corresponds two normals: one with $\vecn=\vecnu$ on $D_g^+$ and the other with $\vecn=-\vecnu$ on $D_g^-$. From eq. (\ref{dXdVk}) we have:
%\begin{equation}
%   \frac{\partial \vecX}{\partial V_k}.\vecnu = \frac{\partial \vecx}{\partial V_k}.\vecnu
%\end{equation}
%Then, as $\frac{\partial \vecx}{\partial V_k}$ depends only upon the curvilinear abscissa $s$, proves that the first term of (\ref{demomarc}) equals to zero:
%\begin{equation}
%   \int_{\partial D_g}   \vecn.\frac{\partial \vecX}{\partial V_k}\quad \d l =
%   \int_{\partial D_g^+} \frac{\partial \vecx}{\partial V_k}.\vecnu \d l -
%   \int_{\partial D_g^-} \frac{\partial \vecx}{\partial V_k}.\vecnu \d l
%\end{equation}
%
\begin{equation}
    \iint_{D_g} \left( f-g \right)\left( \grad(g).\frac{\partial \vecX}{\partial V_k} \right) \d S = 0.\label{dphidVk}
\end{equation}
The next step consists in considering the Taylor expansion of $g$ up to the first order:
\begin{equation}
    g(V_k+\Delta V_k) = g(V_k) + \grad(g).\frac{\partial \vecX}{\partial V_p} \Delta V_p\label{varg}
\end{equation}
which, when introduced in Eq.~(\ref{dphidVk}), gives:
\begin{eqnarray}
    && \Delta V_p \iint_{D_g} \left( \grad(g).\frac{\partial \vecX}{\partial V_k} \right) \left( \grad(g).\frac{\partial \vecX}{\partial V_p} \right) \d S\nonumber\\
    &=& \iint_{D_g} \left( \grad(g).\frac{\partial \vecX}{\partial V_k} \right) (f-g) \d S\label{eqndic},
\end{eqnarray}
This can be written as a matrix equation:
\begin{equation}
    M_{kp}\Delta V_p = L_k\label{Mkp},
\end{equation}
%
% in which $M_{kp}$ and $L_k$ are known quantities dependent only on $f$, $g$ and $\partial{\vecX}/\partial{V_k}$. 
This equation represents a simple linear square matrix problem ($N+4$ dimension); its solution $\Delta V_p$ is used to update the shape of the virtual beam. The iterative process is repeated until the value of $\Phi$ decreases by less than a prescribed amount ($10^{-6}$) between two steps. 

The term $\grad(g).\partial \vecX/\partial V_k$ is involved in both the expressions of $M_{kp}$ and $L_k$. 
From Eq.~(\ref{graylevel}) and the beam geometry follows:
\begin{equation}
    \grad(g)=l'(r) \vecnu \label{gradg},
\end{equation}
and, from Eq.~(\ref{defX}):
\begin{equation}
    \frac{\partial \vecX}{\partial V_k} =
    \frac{\partial \vecx}{\partial V_k} - r \frac{\partial\theta}{\partial V_k}
    \vectau\label{dXdVk}.
\end{equation}
The second term, collinear to $\vectau$, does not need to be computed as it is orthogonal to $\grad(g)$.
%
% d theta / d VK dont on n'a pas besoin...
%From Eq.~(\ref{thetaa}), setting $\vecx_0=x_{0,1} \vec{e_1} + x_{0,2} \vec{e_2}$, the derivatives $\partial\theta / \partial V_k$ write:
%
%\begin{equation}
%    \frac{\partial \theta}{\partial x_{0,1}} = 0,
%    \frac{\partial \theta}{\partial x_{0,2}} = 0,
%    \frac{\partial \theta}{\partial \theta_0} = 1,
%    \frac{\partial \theta}{\partial A_n} =
%    L\tilde{\theta}_n\label{dthetadAn},
%\end{equation}
%
The derivatives of $\partial \vecx/\partial V_k$ are obtained using Eq.~(\ref{defx}) and (\ref{thetaa}):
\begin{equation}
    \frac{\partial \vecx}{\partial x_{0,1}} = \ex,
    \frac{\partial \vecx}{\partial x_{0,2}} = \ey,
    \frac{\partial \vecx}{\partial \theta_0} = \int_0^{s} \vecnu \d \xi,
    \frac{\partial \vecx}{\partial A_n} = L \int_0^{s} \tilde{\theta}_n \vecnu \d \xi\label{dzdAn}.
\end{equation}
\begin{figure}[htbp]
\begin{center}
\includegraphics[scale=0.5]{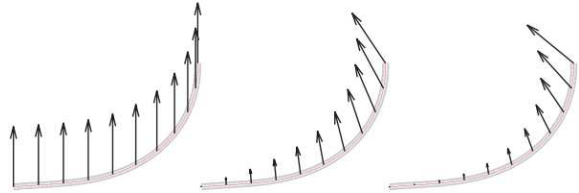}
\caption{Examples of unitary displacement fields. From left to right: $\partial \vecx/\partial x_{0,2}$ (vertical translation), $\partial
\vecx/\partial \theta_0$ (rotation) and $\partial \vecx/\partial A_0$ (uniform increase of the curvature).}
\label{champs_de_deplacement}
\end{center}
\end{figure}
%
% Leur calcul est rapide
% The simple expression of both $\grad(g)$ and $\partial \vecX/\partial V_k$ allows fast computation.
% The fields $\grad(g)$ and $\partial \vecX/\partial V_k$ are analytically defined (except the simple numeric integral in Eq.~(\ref{dzdAn})) and can be computed efficiently.
% les dXdVk sont des champs cinématiques
The displacement of the virtual beam between two steps is close to $(\partial \vecX/\partial V_k) \Delta V_k$ where the fields $\partial \vecX/\partial V_k$ represent unitary kinematic fields.
Figure \ref{champs_de_deplacement} shows an example of such fields (for clarity the fields are only represented at the mean line location, \emph{i.e.} $\partial \vecx/\partial V_k$).
% Idem Correli
% Independence des champs ON LE MET PAS ? PAS DE PREUVE EXACTE..
They play the same role as the unitary displacement fields used in the DIC method \cite{Hild2006,Hild2006b,Rethore2007}.

% Les maillages
The virtual image G is naturally discretized over the curvilinear frame $(s,r)$ that does not correspond to the square grid of image F.
% with a curvilinear abscissa increment $\delta S$ and a radius increment $\delta R$. The corresponding points does not correspond, in general, to the pixel grid of F. 
To avoid any loss of information, the mesh size of the virtual image is much smaller. 
% chosen thinner than the pixel size: $(\delta S<1, \delta L<1)$.
The computation of the second member of Eq.~(\ref{eqndic}) requires one to project the luminance field $f$ onto the mesh of G using, here, a cubic interpolation.

% La longueur
The overall length $L$ of the beam is currently not straightforwardly obtained by the VIC method. 
The end of the fiber is detected from the sharp variation of $\partial \Phi / \partial s$ which occurs at the point where the virtual beam exceeds the end of the fiber.
\section{Determination of the initial conditions}
\label{sec:init}

% Pb des conditions initiales
It has been shown that the VIC method, detailed in the previous section, requires that the boundary of the virtual beam initially surrounds the image of the physical fiber. This section briefly describes the method used to determine these initial conditions.
% Funambule
% The approximative shape of the beam is obtained by a simple algorithm in which the beam is discretized in a set of straight segments (typically some tenth, depending upon the curliness of the object).

The beam is first approximated by a sequence of straight segments. Their positions are given by a simplified VIC method in which the kinematic field is a single rotation around the first point of the segment.
% Each segment orientation is one by one defined by image correlation. This algorithm can be seen as a simplified DIC method, in which each segment has only one kinematic field (the rotation). Let us remark that this approximative shape could be obtained by another method, for example skeletonizing.
%
% Récupération de l'information
Finally the approximate shape is defined upon a collection of equally spaced points $\hat{\vecx}_q$ (with $0\leqslant q \leqslant Q$). This gives the segments angles $\hat{\theta}(\hat{s}_q)$ where $\hat{s}_q$ are the curvilinear abscissae of points $\hat{\vecx}_q$. 
% Determination de Vk0
These data are used to define the set of initial parameters $V_k^0 = \{\hat{\vecx}_0,\hat{\theta}_0,\hat{A}_n\}$ that will be used in the first step of the VIC computation. The first term, $\hat{\vecx}_0$, corresponds to the user-defined position of the initial point.
The other terms are obtained by setting $A_{-1}=\theta_0 / L$ and $\tilde{\theta}_{-1}(\tilde{s})=1$ in Eq.~(\ref{thetaa}) which becomes:
% The other terms are obtained from Eq.~(\ref{thetaa}): setting $A_{-1}=\theta_0 / L$ and $\tilde{\theta}_{-1}(\tilde{s})=1$, the latter can be rewritten as:
%
\begin{equation}
    \frac{\theta(s)}{L} = \sum_{n=-1}^{N} A_n \tilde{\theta}_n(\tilde{s}).
\end{equation}
In particular, this equation applies to each value $\hat{\theta}(\hat{s}_q)$ obtained from the previous approximative analysis, leading to:
\begin{equation}
	\frac{\hat{\theta}(\hat{s}_q)}{\hat{L}} = \sum_{n=-1}^{N} \hat{A}_n \tilde{\theta}_n\left(\frac{\hat{s}_q}{\hat{L}}\right),
\end{equation}
where $\hat{L}=\hat{s}_{Q+1}$. From Eq.~\ref{defthetan}, the functions $\tilde{\theta}_n$ depend only on the series functions $\tilde{\gamma}_n$. Setting the order $N$ of the series such as $N+2=Q+1$ (the number of points $\hat{\vecx}_q$), this equation represents a linear square matrix system whose resolution gives the terms $\hat{A}_n$ (where $\hat{A}_{-1} = \hat{\theta}_0 / \hat{L}$).
% As soon as the order $N$ of the series is chosen such as $N+2=Q+1$ (the number of points $\hat{\vecx}_q$), and the basis functions $\tilde{\theta}_n$ is chosen, this equation represents a simple linear square matrix system whose resolution gives the terms $A_n$ and $\theta_0 = LA_{-1}$.
This defines $V_k^0$ and the VIC computation may start at this order $(Q-1)$, or at a lower one if $V_k^0$ is truncated.
% (the last term generally contain high frequency information that is only relevant of the imprecision of the approximately defined points).

%---------------------------------------------------------------------
\section{Example of applications of the VIC method}
\label{sec:exemples}

% Intro de la section
This section describes two examples especially chosen to illustrate, from a practical point of view, the different steps involved in the technique. The first example (Sec.~\ref{sec:beam}) validates the accuracy of the method for a simple geometry; the second (Sec.~\ref{sec:fiber}) demonstrates its robustness in the case of a curled shape and a low quality image.

% - - - - - - - - - - - - - - - - - - - - - - - - - - - - - - - - - - - 
\subsection{The cantilever beam}
\label{sec:beam}

\begin{figure}[htbp]
\begin{center}
\includegraphics[height=2in]{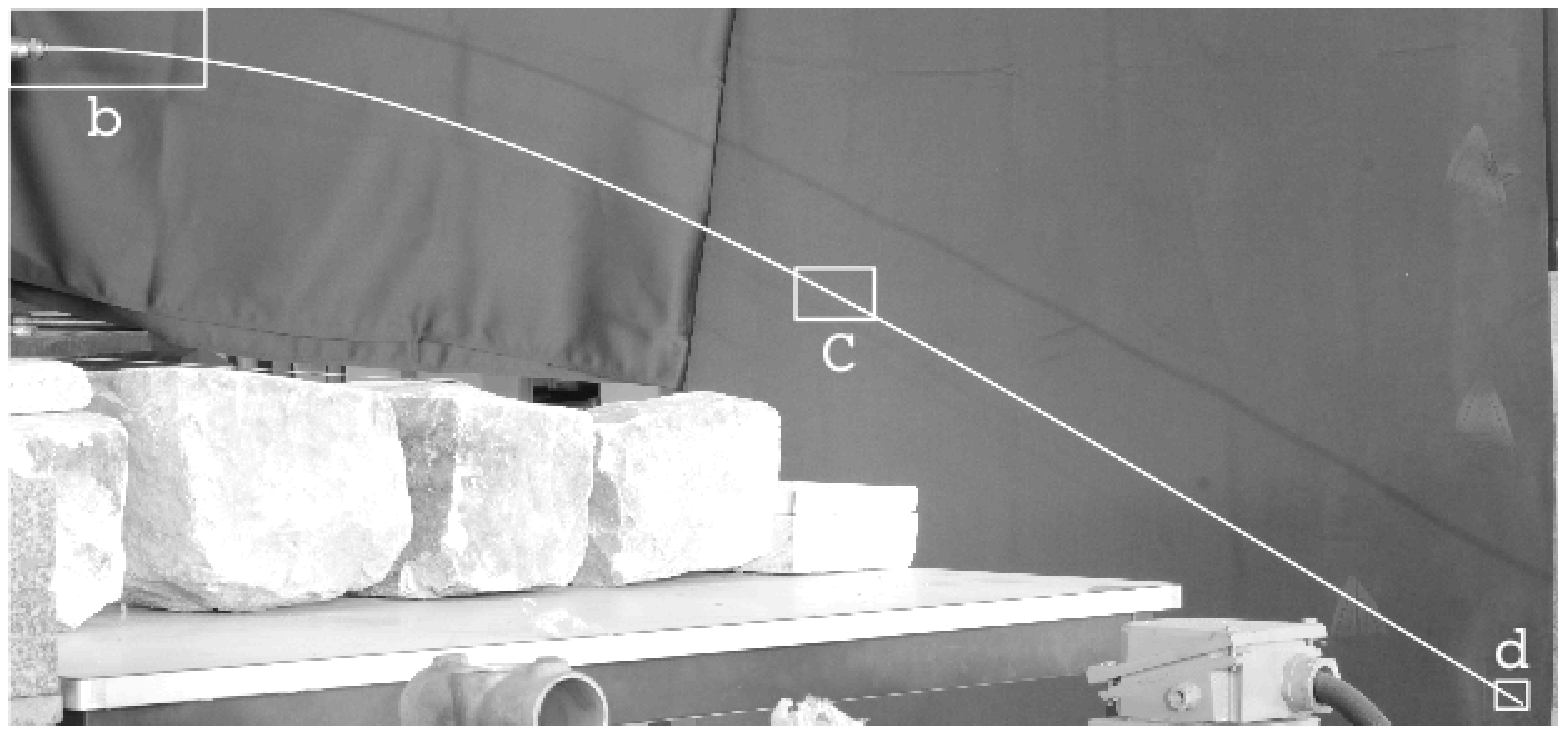}(a)
\includegraphics[height=1in]{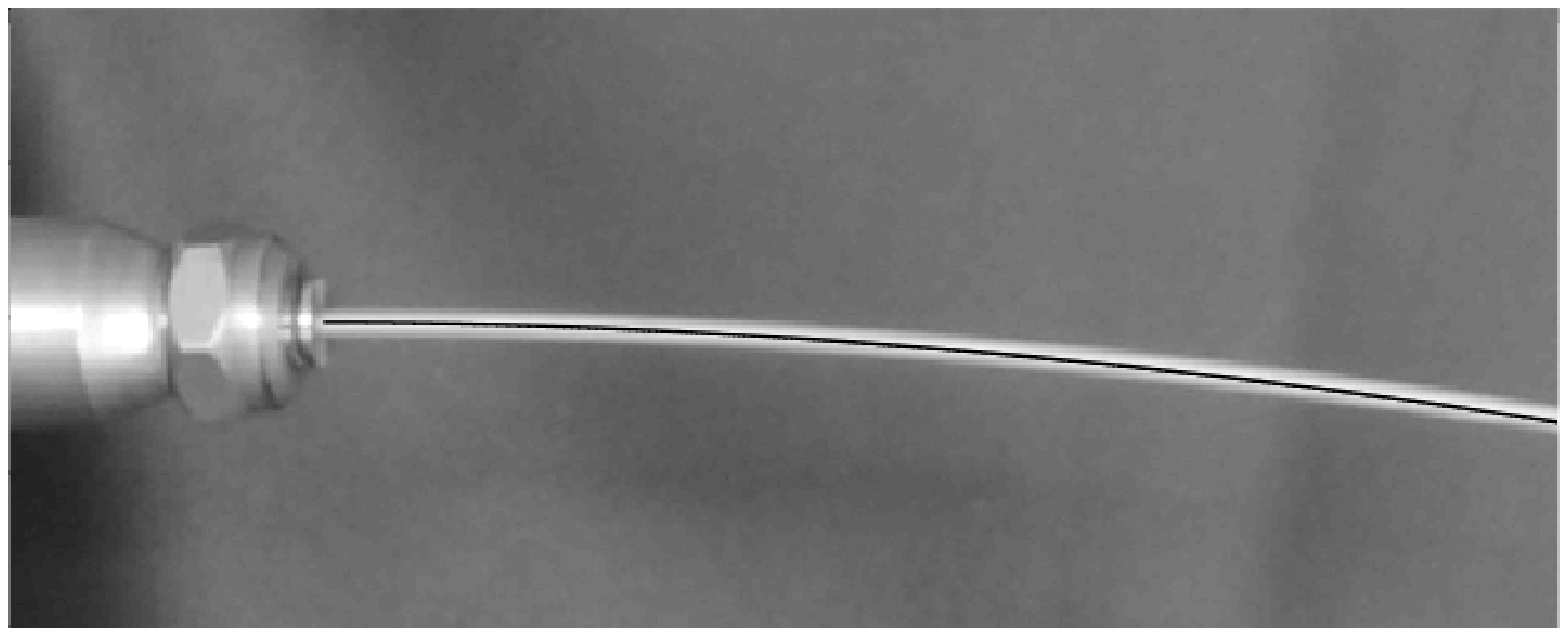}(b)
\includegraphics[height=1in]{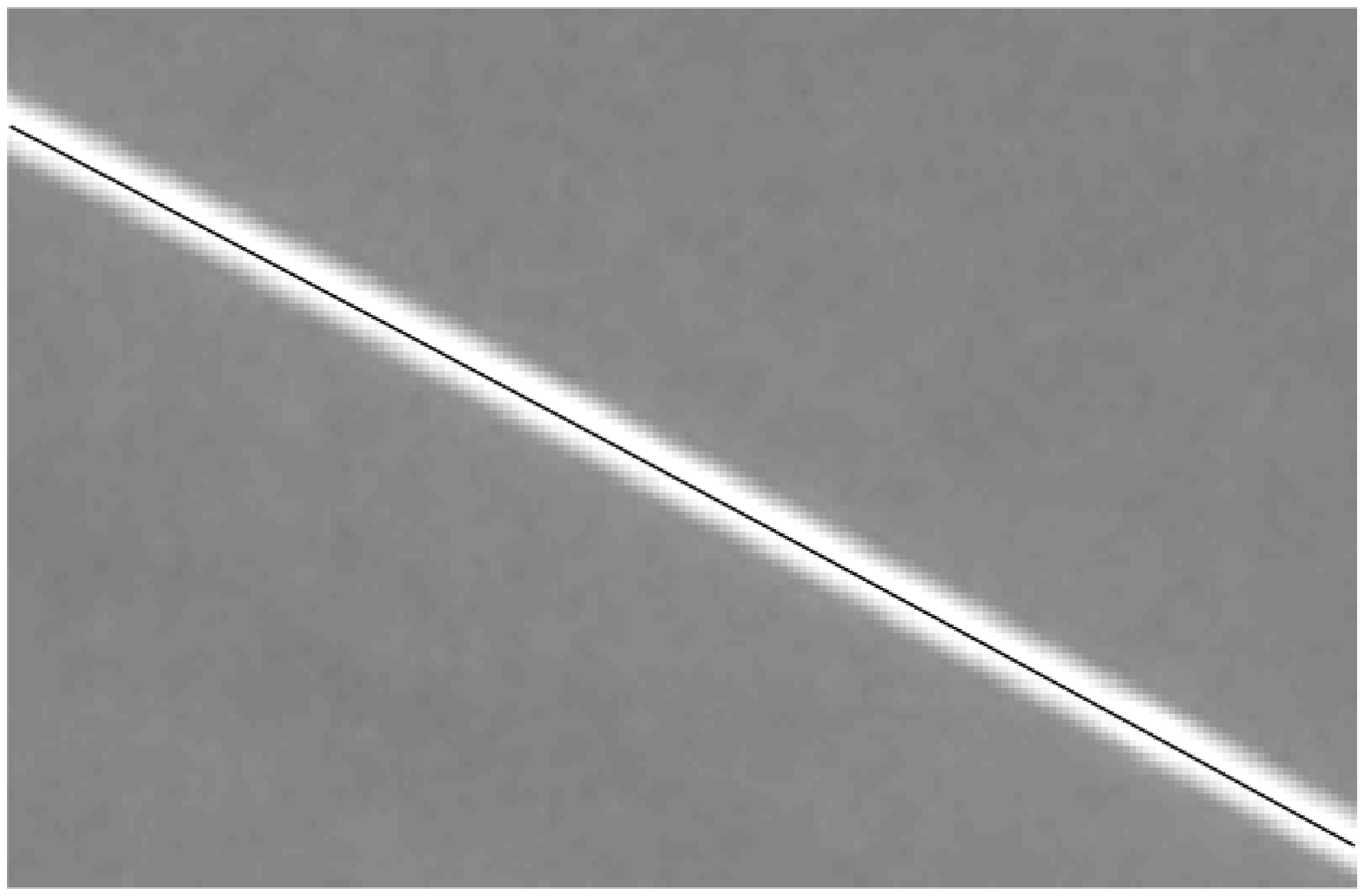}(c)
\includegraphics[height=1in]{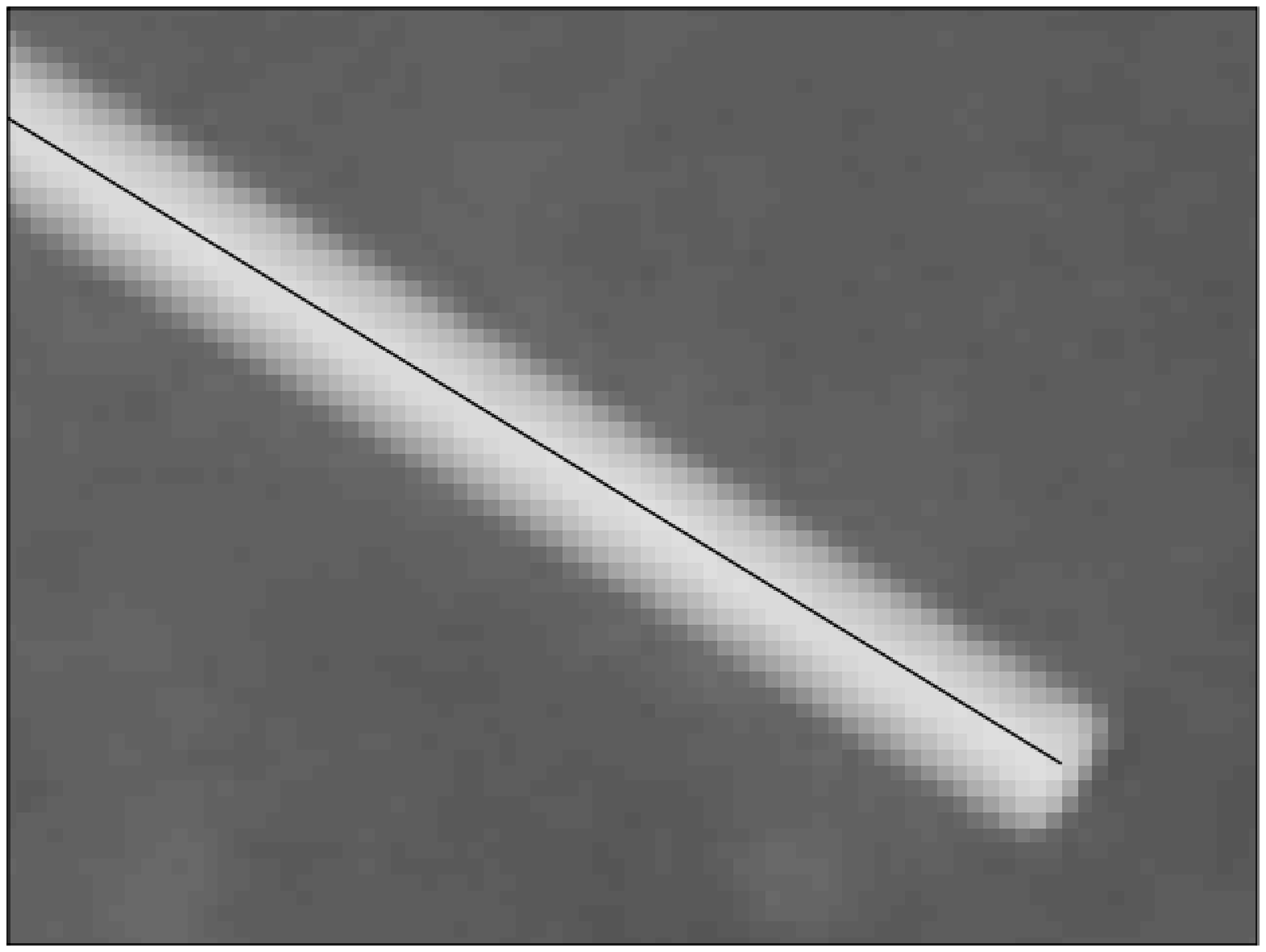}(d)
\caption{(a) Aluminium bar bending under its own weight and identified mean line (white).
(b-d) Magnified views of regions (b) close to the chuck, (c) midway along the bar and (d) at the tip. Solid black line: mean line obtained by the VIC method for a Legendre series of order 3.} \label{alu}
\end{center}
\end{figure}

% Intro
% In the present section we compare the theoretical result of a cantilever bar bending under its own weight to the measurement provided by the VIC method.
% Compared to other existing methods, our technique has the advantage of providing an analytical expression for the mean line.
% This is very important for determining quantitatively the mechanical characteristic of bars.
% The present section illustrates the ability of our technique to reach this goal. 
We proceeded to a simple experiment of a cantilever (clamped at one end) straight bar, bent under its own weight (Fig.~\ref{alu}a). 
%
% Description de l'expérience
The 2017-T4 aluminium bar has a length of $2459$ mm and a radius of $4.95$ mm. 
It is left free to bend under its own weight with one of its ends clamped in an horizontal chuck (Fig.~\ref{alu}b).
A black curtain was placed in the back of the device in order to obtain a good intensity contrast and an uniform background (the latter condition is detailed in Sec.~\ref{sec:adj}).
A single light has been positioned close to the camera in order to illuminate equally both sides of the rod (in order to have a symmetrical luminance function $f(r)$). 
% (this corresponds to the symmetry of the function $f(r)$ in Annex~\ref{annex1}).
Once the bar is at rest (this may take a few minutes) an high resolution picture is captured using a Nikon D300 digital camera.
This image (cropped from the original one) has $3897 \times 1841$ pixels.
The field of view represents $2.33 \times 1.09$ m$^2$ in the focal plane.

% Choix de la série de Legendre
For this image, we used Legendre series for the functions $\tilde{\gamma}_n$ in Eq.~(\ref{defan}). The Fourier series used in Sec.~\ref{sec:fiber} provides similar results, but an higher order is needed. The Legendre series, in their shifted version ($\tilde{s}\in[0..1]$), can be written as:
\begin{equation}
    \tilde{\gamma}_n = P_{nk}\tilde{s}^k,
    \label{legendre1}
\end{equation}
where $k\in[0,N]$ and with \cite{MathsHandbook}:
\begin{equation}
    P_{nk} = (-1)^{(n+k)}
    \left(\begin{array}{c}
    n \\
    k
    \end{array}\right)
    \left(\begin{array}{c}
    n+k \\
    k
    \end{array}\right),
    \label{legendre2}
\end{equation}
in which the parenthesis refers to combination formulas. 
% xo1 fixé
Yet, for our particular example, because of the small contrast between the chuck maintaining the bar and the bar itself (see Fig.~\ref{alu}b), the abscissa $x_{0,1}$ was fixed so as to avoid any unwanted inclusion of the clamping device onto the virtual beam. The set of shape parameters (introduced in Sec.~\ref{sec:adj}) reduces to $V_k = \{x_{0,2},\theta_0,A_n\}$.
% Ordre 3
% The initial parameters $V_k^0$ obtained from the method described in Sec.~\ref{sec:init}, the VIC method was used while retaining an order $N=3$ for the Legendre series.

% Résultat en images
Fig.~\ref{alu} shows the result of the identification for an order $N=3$: the mean line of the virtual beam perfectly follows the middle of the aluminium bar along its full length.
Due to its structure, the VIC method is not influenced by large and illuminated objects present in the foreground of the picture (like the stones on the bottom left of Fig.~\ref{alu}a).
% Details technique
The virtual beam diameter was set at $20$ pixels (the physical bar diameter is around 10 pixels).
The virtual beam mesh has $61\times8634$ points (approximatively three times thinner than the image grid). The total time required for the computation was about $3$ seconds per iteration using a 2.8 GHz dual-core computer and an algorithm implemented on Matlab.

% Resultats numeriques
% 1.0e+02 *
% Columns 1 through 4
% 1.014116750681612   0.000032806885592   0.000001316739838  -0.000002048839187
% Columns 5 through 6
% 0.000000747375014  -0.000000044683870

% \begin{table}[h!]
%\caption{Value of the parameters defining the position of the mean line of the virtual beam obtained by the VIC method. The position is defined over a Legendre basis defined up to order $N=3$. $x_{0,1}$, $x_{0,2}$ are the coordinates of the fixed end. The angle at that position is $\theta_0$ and $A_n$'s are the coefficient associated to the Legendre polynomials. $L$ is the length of the bar determined by the VIC method.}
%\begin{center}
%\begin{tabular}{|c|c|c|c|c|c|c|c|}
%\hline
%$x_{0,1}$ (pixels) & $x_{0,2}$ (pixels) & $\theta_0$ (deg) & $A_0$ & $A_1$ & $A_2$ & $A_3$ & $L$ (mm)\\
%\hline
%$101.92$ & $101.41$ & $0.188$ & $1.317\,10^{-4}$ & $-2.049\,10^{-4}$ & $0.747\,10^{-4}$ & $-0.045\,10^{-4}$ & $2264$\\
%\hline
%\end{tabular}
%\end{center}
%\label{tabres}
%\end{table}%
%\subsubsection{Quantitative comparison between the VIC and the beam theory}
%
We now compare these results to the prediction of the elastic beam theory \cite{Timoschenko1924}. According to it, the flexural moment $M(s)$, given by static equilibrium, is proportional to the local curvature:
\begin{eqnarray}
	M(s) &=& \int_{s}^{L} \rho \left(x_1(\xi)-x_1(s)\right) \d \xi,\\
    M(s) &=& E\frac{\pi R^4}{4}\gamma(s) \label{eq:momentum},\\
    \theta(0) &=& 0,
\end{eqnarray}
where the abscissa $x_1$ is measured along the horizontal axis. This system is solved using a numerical iterative method and the physical properties $\rho=2700$ g/cm$^3$ and $E=72$ GPa \cite{lemaitre_88}. The Young modulus was confirmed by a three-point bending test performed on the actual specimen, leading to $72.6$ GPa. 
% This result will now be compared to the solution given by the VIC method from the picture of the experiment.

%
\begin{figure}[htbp]
\begin{center}
\includegraphics[scale=0.12]{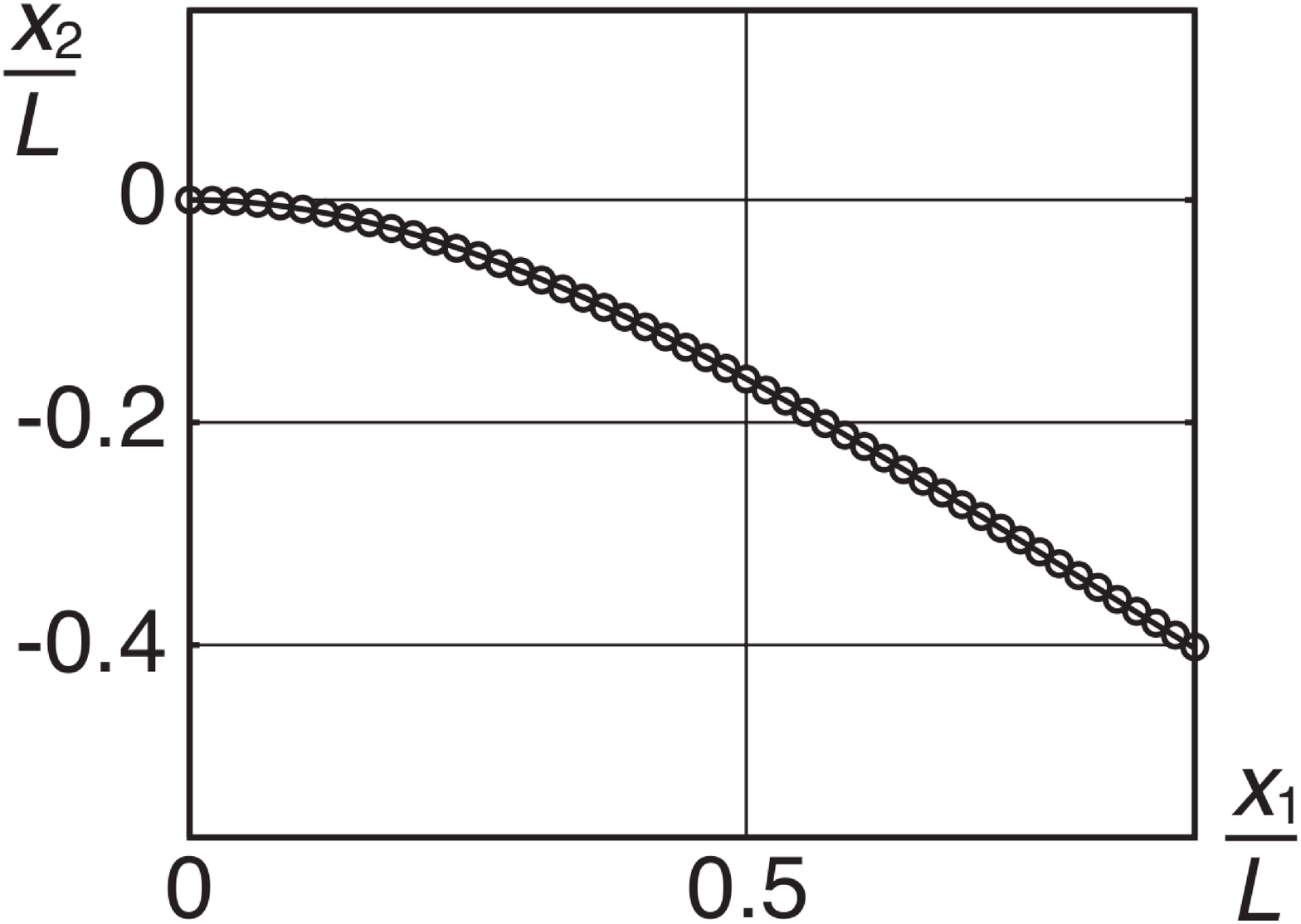}
\includegraphics[scale=0.12]{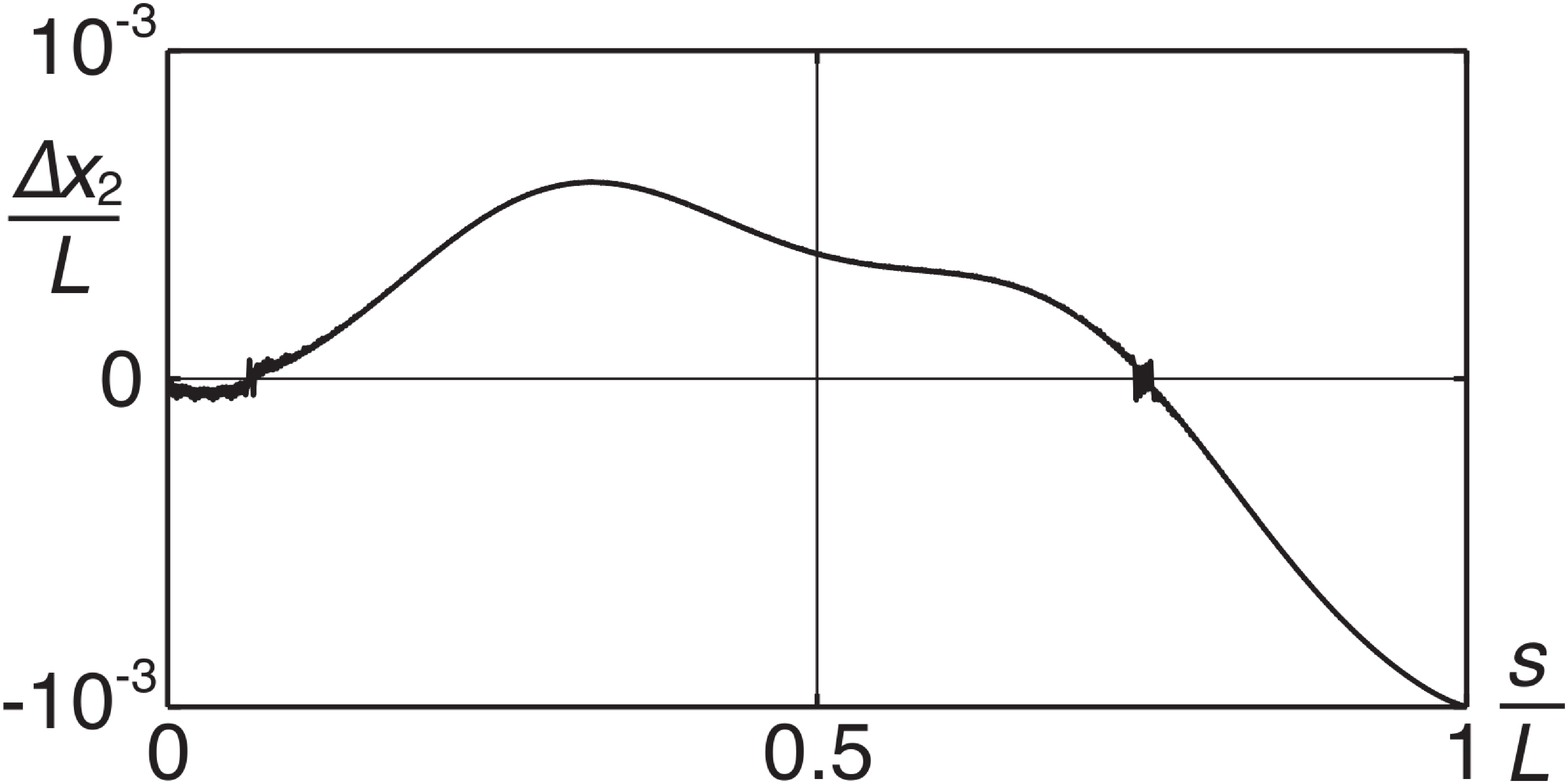}\\
(a) \hspace{6cm} (b)\\
\includegraphics[scale=0.12]{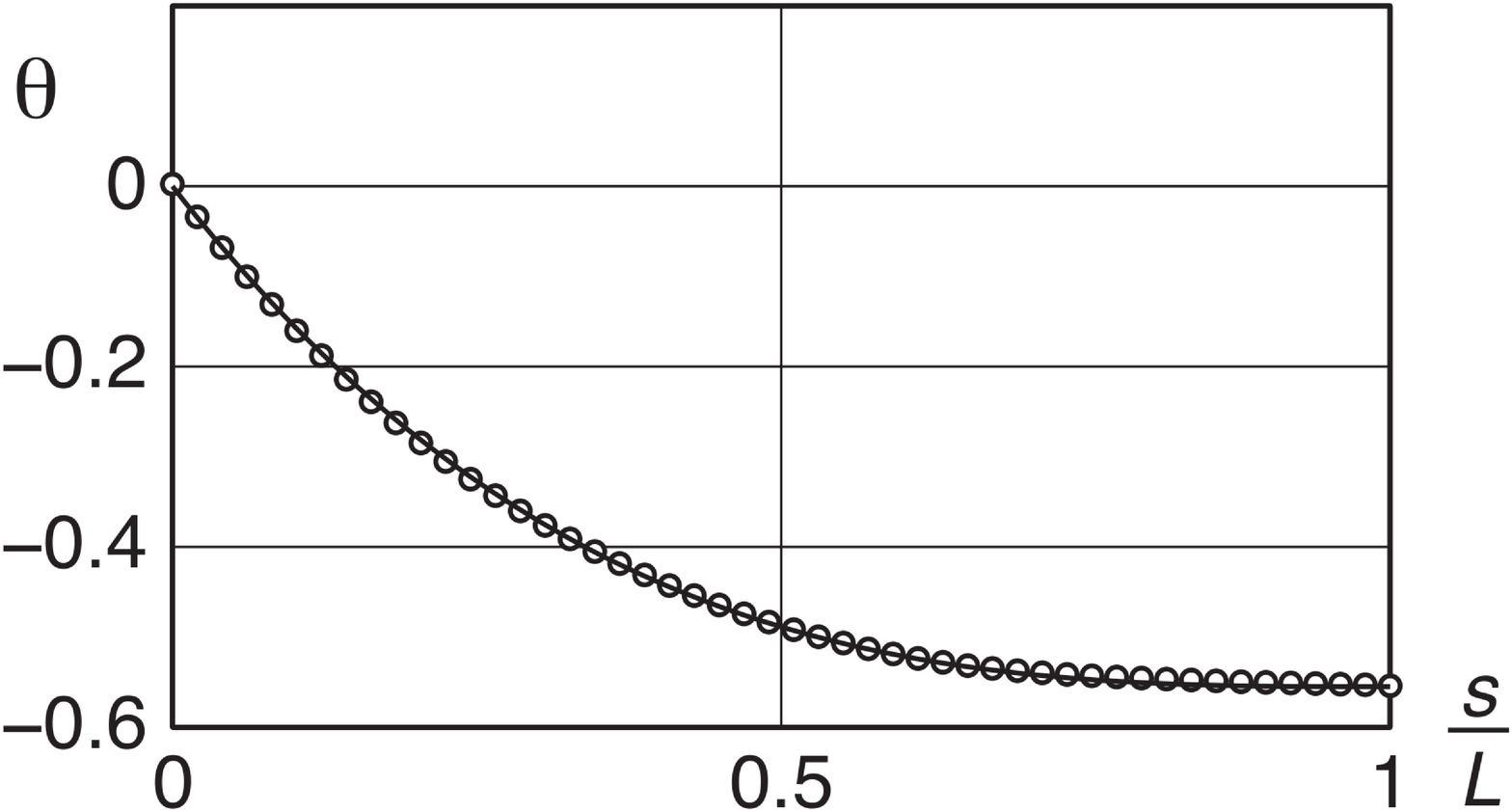}
\includegraphics[scale=0.12]{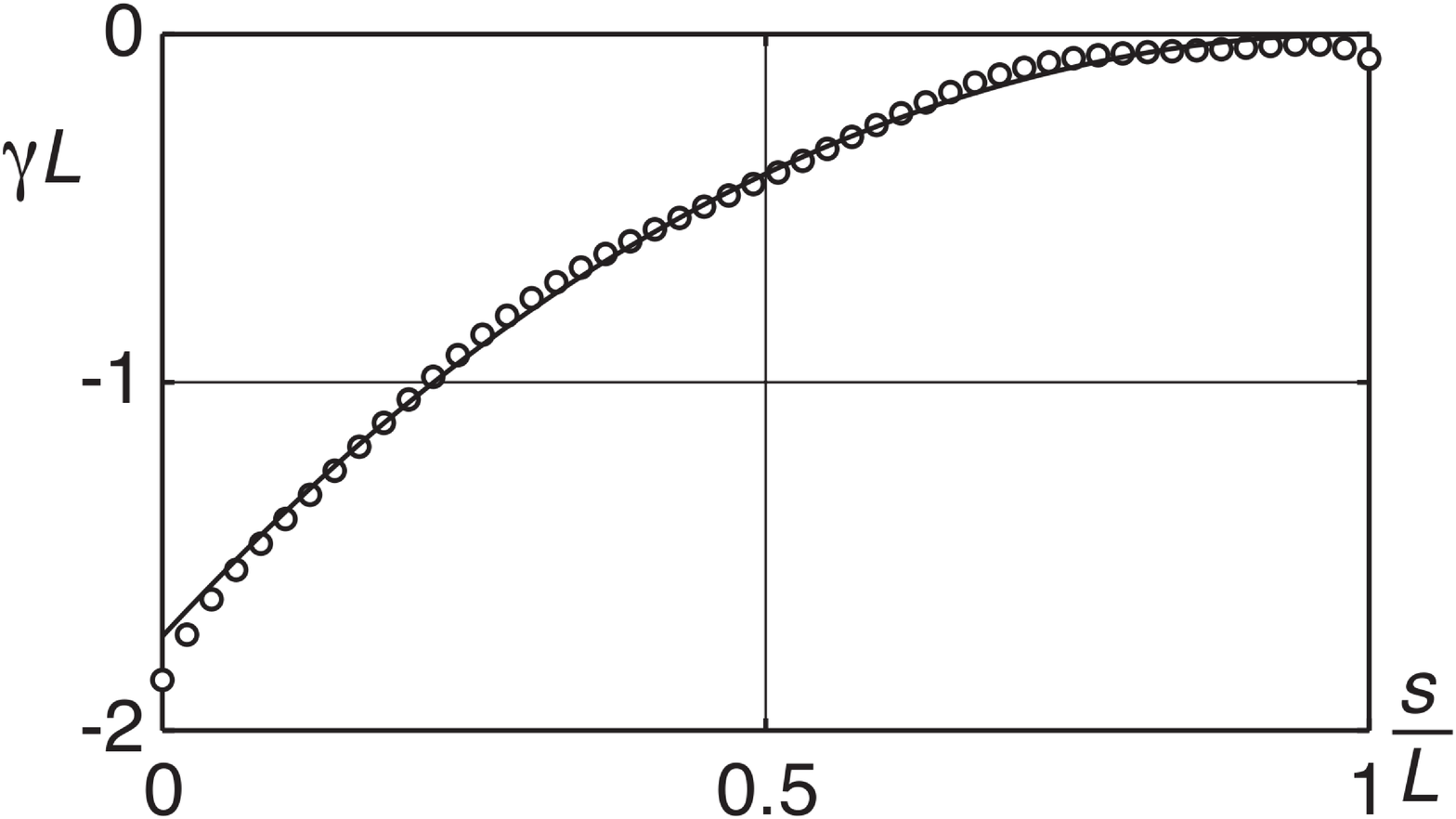}\\
(c) \hspace{6cm} (d)\\
\caption{Comparison between the beam theory (solid line) and the VIC method (circles). (a) Displacement. (b) Discrepancy between the two determinations of the vertical displacement. (c) Angle (radians). (d) Curvatures.}\label{curvature}
\end{center}
\end{figure}
%
% coordinates -normalized by the bar length- of the mean line of the bar obtained by the VIC method (circles) and by the beam theory (solid line). (b) difference between the position inferred from the VIC method and the beam theory as function of the normalized curvilinear abscissa $s/L$. (c) and (d) show the variation of the angle $\theta$ (in radian) and of the normalized curvature $\gamma L$ as function of $s/L$.
% Comparaison VIC/Timoshenko

Fig.~\ref{curvature}a shows that the results given by the beam theory match very well the measurements from the VIC method, in term of displacement in the $x_1$ (horizontal) $x_2$ (vertical) plane.
Fig.~\ref{curvature}b presents a magnified view of the same result and only small differences are observed: the mean quadratic vertical distance between the two solutions is approximatively of $2$ pixels (\textit{i.e.} $1.2$ mm).
A similar discrepancy has been observed in a second experiment in which the chuck was rotated by an half turn.
Figs.~\ref{curvature}c and \ref{curvature}d demonstrate that the VIC method determines correctly the slopes and curvatures. In particular, both the initial slope $\theta(s=0)=0$ and the final curvature $\gamma(s=L)=0$ are correctly determined (note that their value were not imposed by the Legendre polynomial sequence used).

% Choix de N par Phi(N)
\begin{figure}[htbp]
\begin{center}
\includegraphics[scale=0.222]{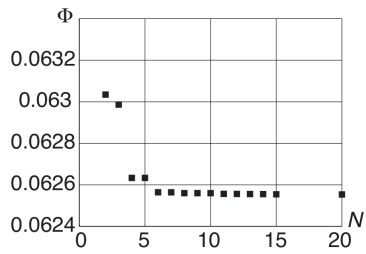}
\caption{Dependence of the final value of the correlation function $\Phi$ on the order $N$.} \label{phideN}
\end{center}
\end{figure}
The choice of the order $N$ results from a classical compromise between simplicity and accuracy. 
% The method used for the determination of the preliminary approximate shape (Sec~\ref{sec:init}) lead to an initial 
% The initial condition determinations (Sec~\ref{sec:init}) gives an initial value $N=Q-1$. 
An order too low (here $N=1$) may lead to a loss of correlation between F and G (the boundary $\partial D_g$ does not enclose anymore the image of the object). Increasing the order monotonically decreases the value of the correlation function (Fig.~\ref{phideN}). When the order becomes too high (for $N>12$ in this example) the matrix $M_{kp}$ in Eq.~(\ref{Mkp}) becomes ill-defined while the level of $\Phi$ asymptotically reaches a minimum value.
%

% Choix de N par image dewrappée
\begin{figure}[htbp]
\begin{center}
\includegraphics[scale=0.333]{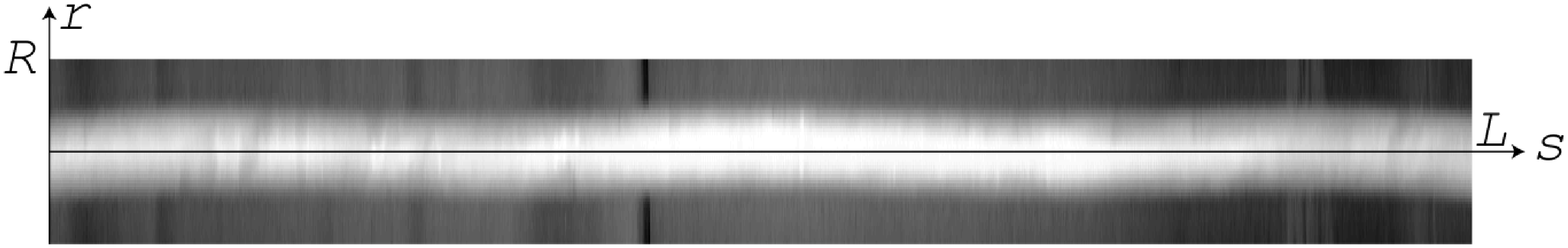}(a)\\
\includegraphics[scale=0.333]{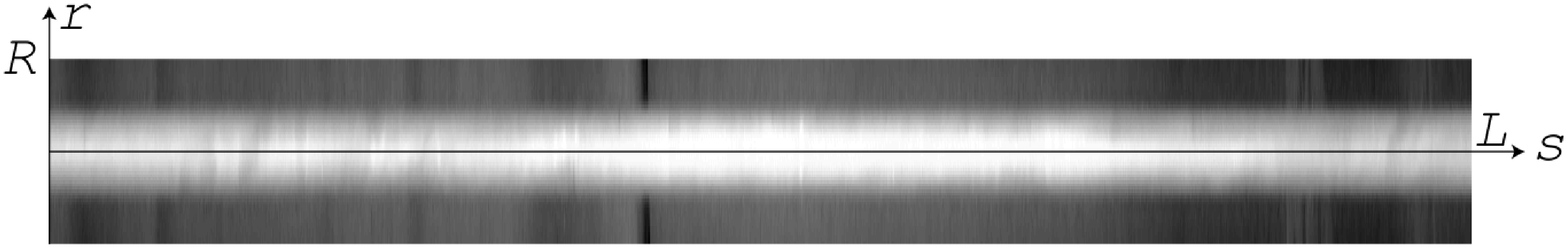}(b)%\includegraphics[scale=0.333]{correl_virtuel.pdf}(c)
\caption{a) and b) : the physical beam image in the unwrapped coordinates of the virtual beam for $N=3$ and $N=8$.}%  c) the virtual beam image in the same frame
\label{alu_f1_corr}
\end{center}
\end{figure}
A visual comparison of the results obtained for two different values of $N$ is displayed in Figs.~\ref{alu_f1_corr}a and \ref{alu_f1_corr}b.
They represent the physical image in the reference frame $(s,r)$ of the virtual beam (represented unwrapped for clarity).
In the case of an ideal correlation, whatever the shape of the fiber, this figure should be symmetrical with respect to the (straight) mean line.
Because of the magnification along the $r$ axis, one can see that the image in Fig.~\ref{alu_f1_corr}a obtained for a low order $N=3$ is slightly wavy while the image in Fig.~\ref{alu_f1_corr}b obtained for $N=8$ is perfectly symmetrical. 
The inhomogeneities of the illumination and of the background are visible but do not influence the final result.
%Furthermore the inhomogeneity of the lightening and the background can be also observed. % (in the definition domain $D_g$) 

% - - - - - - - - - - - - - - - - - - - - - - - - - - - - - - - - - - - 
\subsection{The fiber transported by a fluid flow in a fracture}%%%%%% Voir
\label{sec:fiber}

This second example illustrates the robustness of the technique with regard to the quality of the physical image F and to the complex shape of the object. 

\begin{figure}[htbp]
\begin{center}
\includegraphics[scale=0.667]{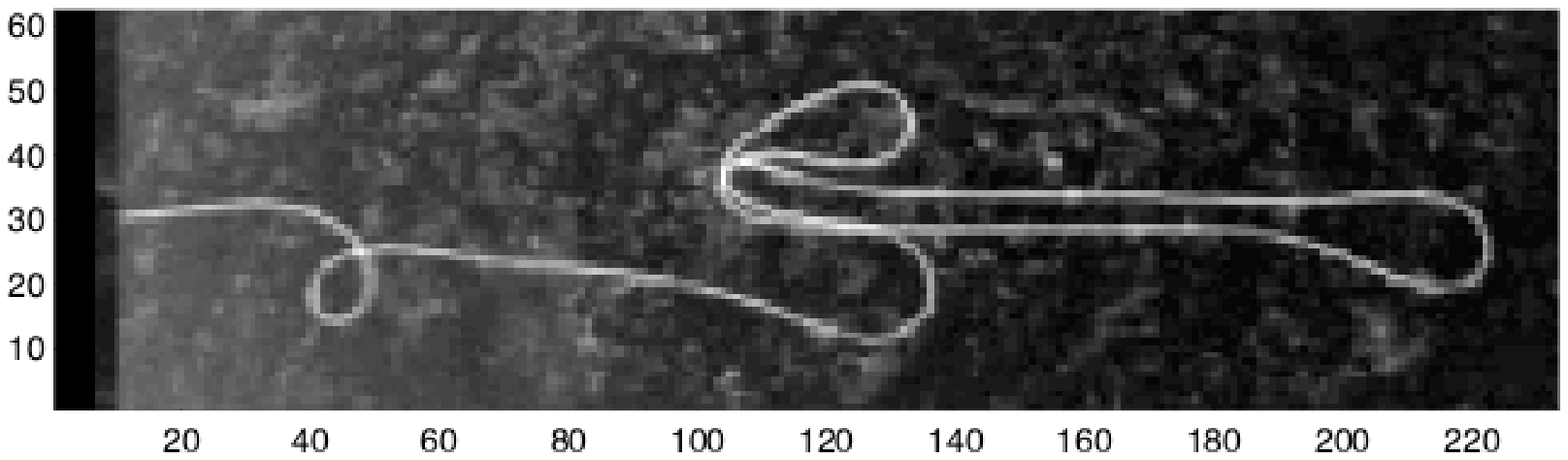}(a)
\includegraphics[scale=0.667]{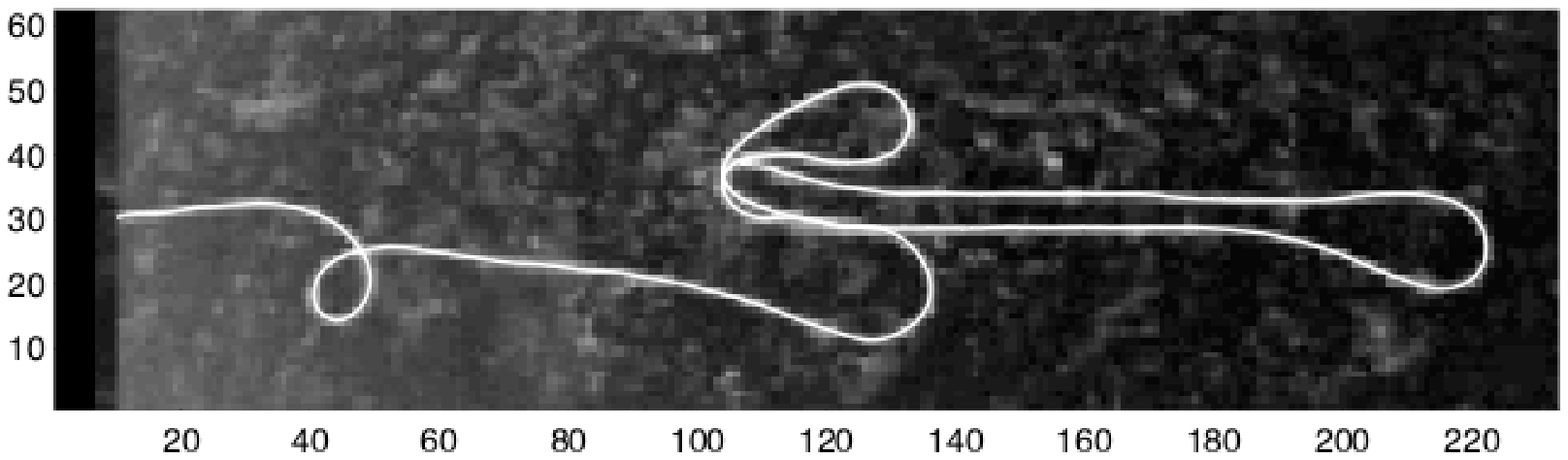}(b)
\includegraphics[scale=0.667]{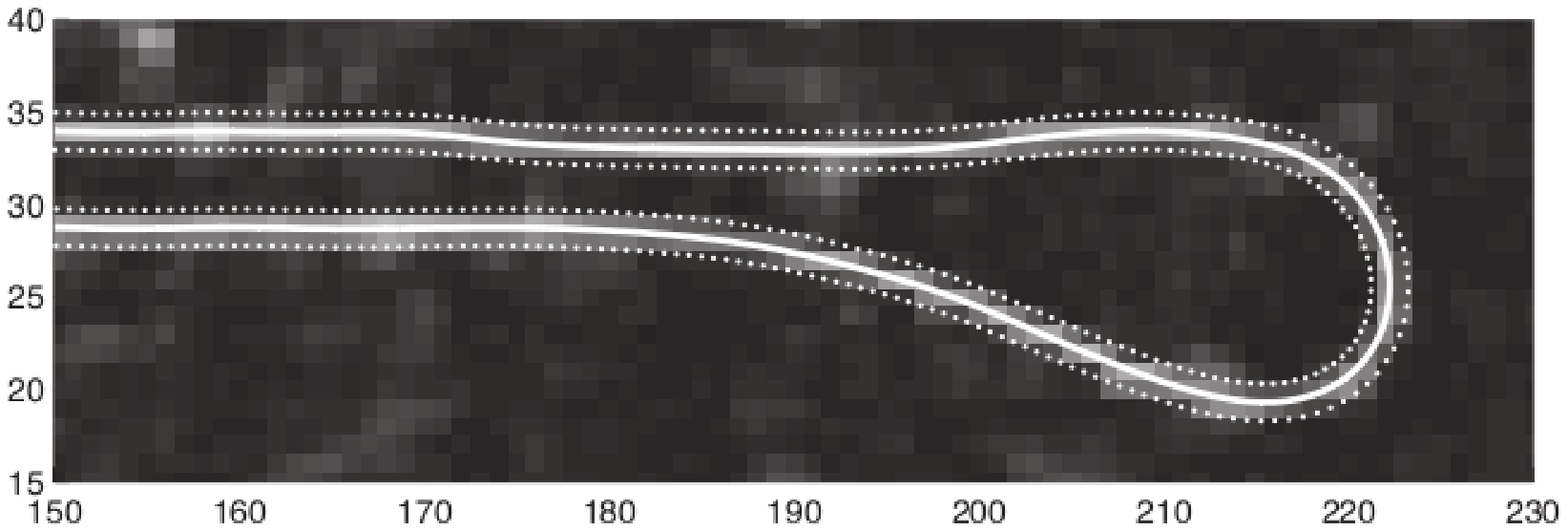}(c)
\includegraphics[scale=0.667]{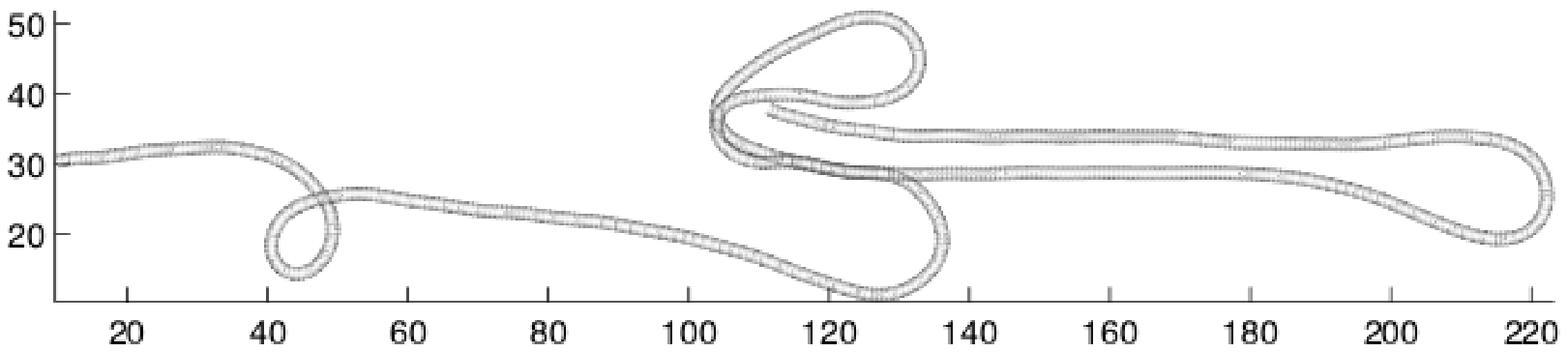}(d)
\caption{
Fiber transported by a fluid flow in a fracture (units are in pixels).
(a) Physical image.
(b) Same image and identified mean line (white line).
(c) Detailed view: identified mean line (white line) and border of the virtual image (white dots).
(d) Virtual beam.
} \label{harold1}
\end{center}
\end{figure}
This example deals with the transport of a fiber by a flow fluid in a transparent fracture made of two rough surfaces \cite{Dangelo2009} (Fig.~\ref{harold1}).
%A flow is established in a fracture with rough walls. The roughness of the walls mimics the roughness of natural fractures and the two halves are placed in front of each others so that a free space of average value $0.75$ mm is left between them. 
Because of the fracture roughness, the free space has a complex geometry leading to a disordered flow velocity field.
As a result, when a fiber of diameter (0.3 mm) close to the fracture aperture (0.75 mm) is inserted into the fracture, it is subject to spatially variable forces from the fluid resulting in strong deformations of the thread.
Fiber transport is only possible at large fluid velocities so that images need to be captured at high rates with a poor light intensity contrast. 
% The Fig.~\ref{harold1} shows an example of such experimental observation.
% The picture has been taken just before the pinning of the fiber. This trapping is the result of the building up of intertwined loops around a pinning site (in the present example this site is located in the middle of the picture).

Several attempts using thresholding operations and spline interpolations were made to extract the center line of the fiber but none of them gave a solution well fitted to the whole profile of the fiber at a given time.
This was especially obvious in situations of large deformation with buckling.
The noise level of the picture was a major problem for thresholding methods (the light intensity of spots in the background is close to that of the pixels corresponding to the fiber).
 
%This lack of reliability is a strong shortcoming for the quantitative analyze of the coupling between the fiber and the fluid flow.
The experimental pictures were them reanalyzed using the VIC method.
A Fourier series containing $N=100$ terms was necessary to describe this complex shape (the previously used Legendre series lead to numerical problems due to large values of $P_{nk}$ in Eq.~\ref{legendre2}).
Fig.~\ref{harold1}c shows that the fiber diameter is close to one pixel then a slightly larger width, $R=1$ pixel, has been selected for the virtual beam.
Similarly to previous example, the abscissa $x_{0,1}$ of the first point (on the left of Fig.~\ref{harold1}a) has been fixed in order to avoid unwanted edge effects.

The Fig.~\ref{harold1} clearly shows that the VIC method gives an accurate estimation for the central line of the fiber. Despite the small radius of the object and the high noise level, the central line collapses everywhere onto the fiber. Wide meanders as well as small deviations from a straight line are perfectly followed. Moreover, even the loops of the fiber can be extracted by this method.

%---------------------------------------------------------------------
\section{Conclusion and discussion}

\label{sec:conclusion} 

% CONCLUSION

% ------------ Etat actuel

% It works
The Virtual Image Correlation method allows one to identify precisely the shape of a fiber from its image.
% Precision
The virtual beam fully encloses the physical image of the fiber so that all the information contained in it is used (and not only its brightest pixels); this feature contributes to the precision of the method.
Furthermore, pixels outside of the definition domain are not used: this speeds up the computation and makes it insensitive to possible artefacts in the background.
% Passe les tests + Divers domaines
The method can deal with noisy images and strongly curved fibers; it can already be used in various domains, from biology to mechanical engineering.
% Force d'avoir les courbures
The analytical identification of the curvatures given by the VIC provides informations on the mechanical state of the fiber: it is possible to identify its flexural properties from the knowledge of external forces. Conversely, if the mechanical parameters are known, one can compute (using an inverse problem approach), the 
external forces acting on the fiber.

% ------------ Améliorations a court terme

% Ameliorations experimentales
% From an experimental point of view, better light conditions and corrections of the optical aberrations would increase the precision of the method. 
% Amelioration theoriques
From a theoretical point of view, it would be interesting to include the length of the fiber in the optimization process.
% Autres series
The examples provided in the present paper use either Fourier or Legendre series decompositions for the representation of the mean line. Other series such as the Chebyshev polynomials or other functions used for Computed Aided Design (B{\'e}zier curves, splines...) may also be used.
Following \cite{Grediac2002, Hild2006}, if the method is used for identifying the mechanical properties of the fiber, one should define the main term of the series from the theoretical beam equation.

% ------------ Améliorations a long terme

% 2D + temps
Time sequences will also be considered: the variation of images with time allow one to increase the precision of the analysis and to study the dynamic of the buckling process.
% 3D
Three-dimensional analysis is also envisaged and will use at least two pictures taken from different angles of view.

% Autres applications
The VIC method can also be applied to images of other curvilinear shapes (non fiber objects) in various scientific domains in which the curvature has a major role, for example thermal or diffusion fronts in fluid mechanics, chemistry... It may also be useful in the "pure" image processing field in some case of line or pattern recognition. With a slight modification of the virtual beam luminance definition (a step shape), the method may also be applied to edge detection problems.

\bibliographystyle{unsrt}
\bibliography{peche}

\end{document}